\documentclass[journal]{IEEEtran}
\usepackage[english]{babel}
\usepackage[utf8]{inputenc}
\usepackage{cite}
\usepackage{amsmath,amssymb,amsfonts}
\usepackage{algorithmic}
\usepackage{graphicx}
\usepackage{textcomp}
\usepackage{xcolor}
\usepackage{url}
\usepackage{placeins}
\usepackage{tikz}
\usetikzlibrary{shapes.geometric, arrows, positioning, calc}
\usepackage{fancyvrb}
\usepackage{booktabs}
\usepackage{caption}

\begin{document}


\title{Improving Small-Scale Large Language Models Function Calling for Reasoning Tasks}

\author{
Graziano A. Manduzio, 
Federico A. Galatolo, 
Mario G. C. A. Cimino, 
Enzo Pasquale Scilingo, 
and Lorenzo Cominelli
\thanks{G. A. Manduzio, F. A. Galatolo, M. G. C. A. Cimino, L. Cominelli and E. P. Scilingo are with the Dipartimento di Ingegneria dell'Informazione, Universit\`a di Pisa, 56122 Pisa, Italy 
(e-mail: grazianoalfredo.manduzio@phd.unipi.it), (e-mail: \{federico.galatolo, mario.cimino, lorenzo.cominelli, enzo.scilingo\}@unipi.it).}}

\markboth{}%
{Cognome \MakeLowercase{\textit{et al.}}: Titolo Breve}

\maketitle

\begin{abstract}

Recent advancements in Large Language Models (LLMs) have demonstrated exceptional capabilities in natural language understanding and generation. While these models excel in general complex reasoning tasks, they still face challenges in mathematical problem-solving and logical reasoning. To address these limitations, researchers have explored function calling abilities, allowing LLMs to execute provided functions and utilize their outputs for task completion.
However, concentrating on specific tasks can be very inefficient for large-scale LLMs to be used, because of the expensive cost of training and inference stages they need in terms of computational resources.  
This study introduces a novel framework for training smaller language models in function calling, focusing on specific logical and mathematical reasoning tasks. The approach aims to improve performances of small-scale models for these tasks using function calling, ensuring a high level of accuracy. Our framework employs an agent that, given a problem and a set of callable functions, queries the LLM by injecting a description and examples of the usable functions into the prompt and managing their calls in a step-by-step reasoning chain. This process is used to create a dataset of correct and incorrect reasoning chain chat completions from a large-scale LLM. This dataset is used to train a smaller LLM using Reinforcement Learning from Human Feedback (RLHF), specifically employing the Direct Preference Optimization (DPO) technique.
Experimental results demonstrate how the proposed approach balances the trade-off between model size and performance, improving the ability of function calling for reasoning tasks, in smaller models. 

\end{abstract}


\section{Introduction}

\IEEEPARstart{R}ecent years have seen the rapid development of Large Language Models (LLMs) that have demonstrated exceptional natural language understanding and generation capabilities.
Research has explored the unexpected abilities of LLMs beyond their primary training task of text prediction \cite{nolfi2023unexpected}. 
These models have shown promise in function calling for software APIs \cite{shen2024hugginggpt, schick2024toolformer, parisi2022talm, patil2023gorilla, li2023api}, boosted by the launch of GPT-4 plug-in features \cite{achiam2023gpt}.
Integrated tools include web broswer, translation system, Dialogue State Tracking (DST) \cite{li2024large} and robotics \cite{vemprala2023chatgpt, wang2024lami}.
Furthermore, while LLMs have shown promising results in general complex reasoning benchmarks, they still face challenges in mathematical problem-solving and logical capacities \cite{sun2023survey}. To address these limitations, researchers have proposed various techniques \cite{qiao2022reasoning, huth2004logic}, including the ability of function calling \cite{kim2023llm}, that allows LLMs to execute provided functions and utilize their outputs to assist in task completion. These functions can vary from basic tools like a calculator \cite{cobbe2021training} that performs arithmetic operations to more advanced methods.
%
%
%
However, concentrating on specific tasks that use only a small portion of available APIs underscores the inefficiency of depending solely on large models like GPT-4, which require significant computational resources for both the training and inference stages \cite{brown2020language, radford2018improving, achiam2023gpt, wu2023empirical}.
This situation calls for the creation of smaller, task-specific LLMs that maintain core functionality while reducing operational costs \cite{shen2024hugginggpt, pallagani2024prospects}.
The trend towards smaller models, while promising, introduces new challenges. One significant concern is the increased likelihood of errors or "hallucinations," which can compromise the accuracy of output formatting \cite{yao2023llm, zhang2023siren, ji2023towards}. Given that precise output formatting is essential for developing robust software applications, this issue becomes particularly critical.
To address the drawbacks of oversized LLMs, which incur excessive training and inference costs, we introduce a novel framework to train smaller language models starting from the function calling abilitities of large models, for specific logical and mathematical reasoning tasks.
This framework involves the use of an agent that, given a problem and a set of possible functions useful for its solution, queries a large-scale LLM by injecting function descriptions and examples into the prompt and managing the proper function calls that the model needs to find the solution, all that in a step-by-step reasoning chain.
This procedure is so used for the creation of a dataset with correct and incorrect chat completions. The generated dataset is then used to train a smaller model using a Reinforcement Learning from Human Feedback (RLHF) \cite{christiano2017deep, stiennon2020learning, ziegler2019fine, ouyang2022training} approach, known as Direct Preference Optimization (DPO) \cite{rafailov2023direct}.
We present the methodology tested on two different types of reasoning tasks, First-Order Logic (FOL) and math. To achieve this goal a set of FOL problems were built ad hoc, taking inspiration from the HuggingFace dataset \texttt{\url{SAGI-1/SYMBOLIC_DATA_PLUS_REASONING_DATA_V1}} \cite{sagi2023symbolic}. Examples of mathematical problems were drew directly from the \texttt{GSM8K} \cite{cobbe2021training, cobbe2021gsm8k} dataset. In Section \ref{sec:rel_works} FOL and the DPO are presented. In Section \ref{sec:methods}, the pipeline and methodologies to generate the dataset and train the small-scale model are shown. Finally, in Section \ref{sec:exps} we present experimental results of our framework, where performance of the trained model is compared to the original one.   

\section{Related Works}
\label{sec:rel_works}


\subsection{First-Order Logic (FOL)}

First-Order Logic (FOL), also known as First-Order Predicate Calculus or Predicate Logic, is a formal system that extends propositional logic to include variables, quantifiers, and predicates. This extension allows for greater expressive power in formalizing mathematical statements and reasoning \cite{enderton2001mathematical, mendelson2015introduction}.
The syntax of FOL comprises both logical and non-logical symbols. Logical symbols include connectives (such as negation, conjunction, disjunction, implication, and biconditional), quantifiers (universal and existential), and parentheses. Non-logical symbols consist of constants, variables, function symbols, and predicate symbols. These components work together to create a rich language capable of expressing complex logical relationships and structures.
The semantics of FOL provide a framework for interpreting formulas and determining their truth values \cite{van2012logic}. 
%
%
A key feature of FOL is the use of quantifiers, which allow for statements about all or some elements in the domain \cite{smullyan1995first}. The universal quantifier expresses that a property holds for all elements in the domain, while the existential quantifier expresses that a property holds for at least one element in the domain. These quantifiers significantly enhance the expressive power of FOL compared to propositional logic.
%
%
FOL finds numerous applications across various fields \cite{huth2004logic, fitting2012first}. In mathematics, it is used for formalizing theories and proofs. In computer science, FOL is applied in the specification and verification of software and hardware systems. The field of artificial intelligence utilizes FOL for knowledge representation and automated reasoning. Additionally, linguistics employs FOL in the study of formal semantics of natural languages.


\subsection{FOL and AI}

The field of artificial intelligence (AI) extensively utilizes First-Order Logic (FOL) for knowledge representation and automated reasoning. FOL's expressive power and formal semantics make it an ideal choice for capturing complex knowledge structures and facilitating inference in AI systems \cite{russell2016artificial}. In knowledge representation, FOL allows for the formalization of domain-specific knowledge, enabling AI systems to reason about objects, their properties, and relationships in a structured manner \cite{brachman2004knowledge}. This capability is crucial in expert systems, where domain knowledge is encoded as logical rules and facts, allowing the system to make informed decisions based on logical inference \cite{giarratano1989expert}. 
In the realm of automated reasoning, FOL serves as the foundation for many theorem-proving systems and logical inference engines \cite{robinson2001handbook}. These systems employ techniques such as resolution and unification to derive new knowledge from existing facts and rules, a process fundamental to various AI applications, including planning and decision-making \cite{kowalski1974logic}. Moreover, FOL has been instrumental in the development of answer set programming, a paradigm for declarative problem solving that has found applications in areas such as constraint satisfaction and automated planning \cite{lifschitz2019answer}.
The integration of FOL with probabilistic methods has led to the development of statistical relational learning and probabilistic logic programming, bridging the gap between logical and statistical AI approaches \cite{de2008probabilistic}. This fusion enables AI systems to reason with uncertainty while maintaining the structured representation offered by FOL. Additionally, FOL has played a significant role in the semantic web and ontology engineering, where it is used to define and reason about conceptual models of various domains \cite{staab2013handbook}. 


\subsection{Examples of First-Order Logic Statements}

To better understand the components and structure of First-Order Logic (FOL), let's examine a specific example statement and break down its elements.
Consider the following FOL statement:
\[
\forall x \exists y (P(x) \rightarrow (Q(x,y) \wedge R(y)))
\]
This statement can be read in natural language as: "For every x, there exists a y such that if P(x) is true, then both Q(x,y) and R(y) are true."

The structure of this statement demonstrates several key features of FOL:
\begin{enumerate}
    \item \textbf{Quantification:} The use of both universal ($\forall$) and existential ($\exists$) quantifiers allows us to make statements about all elements or the existence of elements in our domain.
    
    \item \textbf{Variables:} x and y are used to represent arbitrary elements in the domain, allowing for general statements about the relationships between elements.
    
    \item \textbf{Predicates:} P, Q, and R represent properties or relations. P and R are unary predicates (apply to one variable), while Q is a binary predicate (applies to two variables), showing FOL's ability to express different types of relations.
    
    \item \textbf{Logical Structure:} The statement uses implication ($\rightarrow$) and conjunction ($\wedge$) to create a complex logical structure, demonstrating FOL's ability to express intricate logical relationships.
\end{enumerate}
\textbf{Constants in First-Order Logic:}
In addition to variables, FOL also includes constants, which represent specific, named individuals in the domain of discourse. Constants allow us to make statements about particular entities rather than arbitrary ones.
To illustrate the difference between variables and constants, consider the following two FOL statements:
\[
\exists x (\text{Movie}(x) \wedge \text{ActedIn}(y, x))
\]
\[
\exists x (\text{Movie}(x) \wedge \text{ActedIn}(\text{seanconnery}, x))
\]
The first statement uses two variables, x and y. It can be read as: "There exists a movie x such that y acted in x." This statement asserts the existence of a movie and an actor, but doesn't specify who the actor is.
The second statement uses a variable x and a constant \texttt{seanconnery}. It can be read as: "There exists a movie x such that Sean Connery acted in x." This statement is more specific, asserting the existence of a movie in which the particular individual Sean Connery acted.
The use of the constant \texttt{seanconnery} allows us to make a claim about a specific person, whereas the variable y in the first statement could refer to any actor.
The latter and other examples of FOL statements can be found in the HuggingFace dataset \texttt{\url{SAGI-1/SYMBOLIC_DATA_PLUS_REASONING_DATA_V1}} \cite{sagi2023symbolic}.


\subsection{Direct Preference Optimization}

\begin{figure}[htbp]   
    \centering
    \includegraphics[width=1\linewidth]{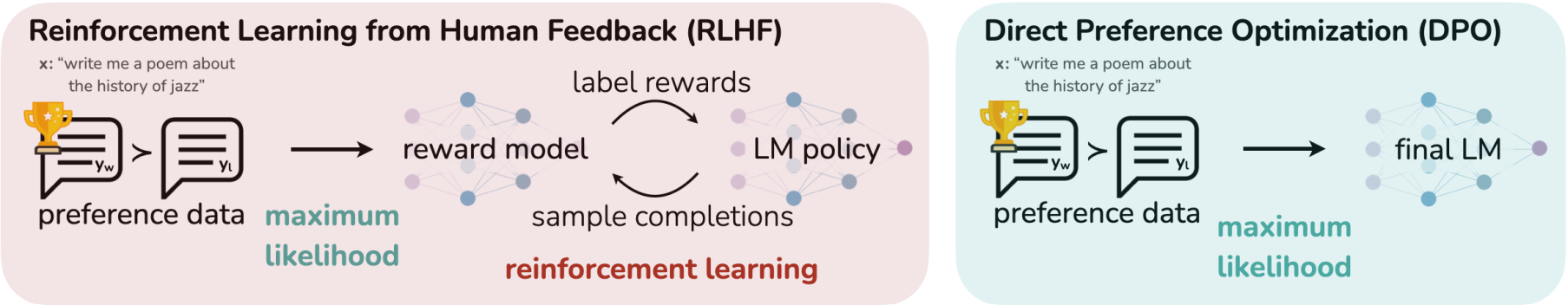}
    \caption{Comparison of typical RLHF approach and Direct Preference Optimization (DPO), \cite{rafailov2023direct}.}
    \label{fig:ppo_vs_dpo}
\end{figure}

While supervised fine-tuning is a common approach, alternative methods leveraging Reinforcement Learning from Human Feedback (RLHF) have gained prominence. One such method is \textit{Proximal Policy Optimization} (PPO) \cite{schulman2017proximal}, which integrates a reward model into the reinforcement learning framework for policy optimization. Despite its effectiveness, PPO's requirement for extensive human feedback to train the reward model makes it resource-intensive and time-consuming.
A more efficient and equally effective alternative is Direct Preference Optimization (DPO) \cite{rafailov2023direct}. DPO distinguishes itself by enabling the model to learn a policy directly from user preference data, eliminating the need for an explicit reward function. Furthermore, DPO has demonstrated superior stability compared to PPO.
The DPO process begins with gathering human feedback. Assessors evaluate pairs of model-generated responses to identical prompts, creating a dataset of preference pairs. Unlike PPO, which trains a separate reward model, DPO incorporates these preferences directly into the training objective. Fig.~\ref{fig:ppo_vs_dpo} illustrates the key differences between these approaches.
DPO's optimization algorithm iteratively adjusts the model parameters to maximize the likelihood of outputs aligning with reviewer preferences. This is achieved through a specialized loss function (Equation \ref{eq:loss_dpo}) that directly penalizes the generation of less-preferred outputs. For a training instance $(x,y_w, y_l)$, comprising an input $x$, a preferred output $y_w$, and a non-preferred output $y_l$, the loss function compares the probabilities of the reference policy $\pi_{ref}$ (the initial policy) with those of the new policy $\pi_{\theta}$ for both preferred and non-preferred outputs.
%
%
\begin{equation}
\begin{split}
\mathcal{L}_{DPO} (&\pi_{\theta}; \pi_{ref}) = \\
& = - \mathbb{E}_{(x, y_w, y_l) \sim \mathcal{D}} \Bigg[ \log \sigma \Big(
\beta \log \frac{\pi_{\theta} (y_w | x) }{\pi_{ref}(y_w| x)} \\
& \quad - \beta \log \frac{\pi_{\theta} (y_l | x) }{\pi_{ref}(y_l| x)} \Big) \Bigg]
\end{split}
\label{eq:loss_dpo}
\end{equation}
This formulation enables DPO to efficiently learn from human preferences, offering a streamlined alternative to traditional RLHF methods.
\section{Methodology}
\label{sec:methods}

\subsection{Proposed framework}

\begin{figure}[htbp]   
    \centering
    \includegraphics[width=1\linewidth]{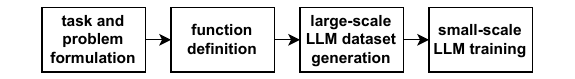}
    \caption{Schematic of the presented framework pipeline.}
    \label{fig:pipeline}
\end{figure}

The pipeline of the presented methodology constitutes of 4 different stages:
\begin{itemize}
\item definition of tasks and problems to be addressed. Calling function abilities of LLMs can be tested on a various corpus of different reasoning tasks, ranging from logical and mathematical to visual and multimodal ones;
\item once tasks and problems are defined, in turn a set of functions needs to be defined for each problem. These functions serve for the LLM to solve the reasoning steps, control the chain flow and verify the intermediate and final responses, working similar to the process-supervised reward models (PRMs) or to the Outcome-supervised Reward Models (ORMs) presented in \cite{lightman2023let};
\item choise of a pre-trained large-scale LLM to generate the dataset of right and wrong completions using a chain-of-thoughts prompting that forces the LLM to reason step-by-step. An agent queries the LLM, stores the response and calls the function until the solution to the problem is obtained. Final solutions can be right or wrong. The chains of thought for each problems are recorded;
\item fine-tuning a small-scale LLM using reinforcement learning on the given dataset. The DPO algorithm is performed.
\end{itemize}
The pipeline of the presented framework is shown in Fig.~\ref{fig:pipeline}.

\subsection{Experimental setup}

To test our framework, the following choices were made:
\begin{itemize}
\item we defined 6 FOL problems, 3 with a single predicate and 3 with double predicates; we then drew 9 mathematical problems from the \texttt{GSM8K} dataset for a total of 15 problems overall;
\item we defined a set of callable function for each problem. For example for the mathematical problems we defined the fundamental operations. Furthermore, we defined the verifier \texttt{CheckCorrectChain()} and the chain flow controller \texttt{Stop()};
\item we created a specific dataset for the DPO algorithm, using Llama3-70B \cite{dubey2024llama} in inference on reasoning tasks, through the Pytorch library \texttt{microchain} \cite{galatolo_microchain_2023};
\item with the new data, we fine-tuned with DPO a smaller model, Mistral-7B-Instruct-v0.2, using a single GPU.
\end{itemize}

Let's see specifically how the dataset was created and how the smaller model was fine-tuned.

\subsection{Microchain}

\begin{figure}[htbp]   
    \centering
    \includegraphics[width=1\linewidth]{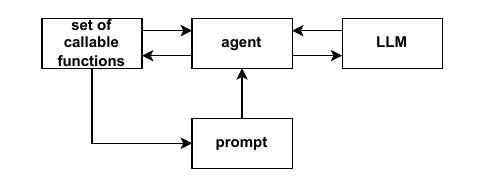}
    \caption{Schematic of the agent-based library \texttt{microchain}, used to generate the dataset.}
    \label{fig:microchain}
\end{figure}

The presented framework generates the dataset using \texttt{microchain}, an agent-based library that given a set of callable functions usable to solve a given problem and coded in a specific python file, enables the agent to query a LLM with a prompt, where these functions are introduced each one with a declaration, a description and a use case example. An instance of prompt $x^i$ sampled from the generated dataset $\mathcal{D}^*$ and related to a FOL problem is shown in Tab.~\ref{tab:movie_expert_prompt}. We see how the chain-of-thought reasoning is explicitly induced by inviting the LLM, that works as a scheduler, to reason step-by-step. After recording the response of the LLM as associated to the \textit{assistant} role, the agent calls the function the LLM needs to solve the reasoning step and proceed to the next one, recording the output as associated to the \textit{user} role and so on, until the final solution is obtained. The \texttt{CheckCorrectChain()} function works as a verifier that checks whether or not the intermediate and the final responses are called in the right order. The same work flow is implemented for a \texttt{GSM8K} dataset problem, but the verifier checks only the correctness of the final solution.
Hence, the agent at each iteration records the entire chain of LLM reponses/function outputs as a set of assistant/user role chat contents, presented in the format commonly used for the chat completions of the LLMs \cite{huggingface2024chattemplating, openai2024chatcompletions}. Therefore the LLM bases its next response on the previous ones recorded in the chain. A block diagram of the \texttt{microchain} library work flow is shown in Fig. \ref{fig:microchain}. 

\subsection{Dataset Creation Process}

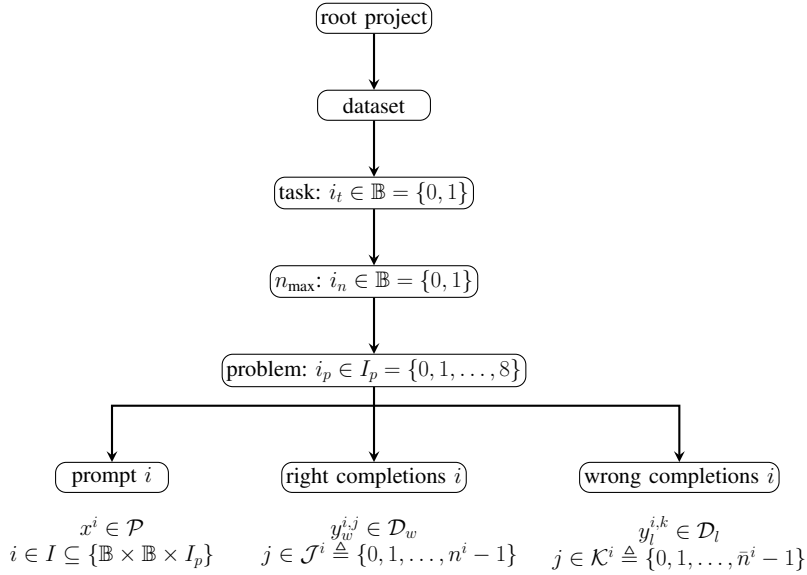
\begin{figure*}
\centering
\begin{tikzpicture}[
    scale=0.5,
    transform shape,
    node distance = 1.5cm and 3cm,
    box/.style = {rectangle, draw, rounded corners, minimum width=3cm, minimum height=0.8cm, font=\LARGE},
    arrow/.style = {->, >=stealth, thick},
    dashed arrow/.style = {->, >=stealth, thick, dashed}
]
\node[box] (root) {root project};
\node[box, below=of root] (dataset) {dataset};
\node[box, below=of dataset] (task) {task: $i_t \in \mathbb{B}=\{0,1\}$};
\node[box, below=of task] (nmax) {$n_{\text{max}}$: $i_n \in \mathbb{B}=\{0,1\}$};
\node[box, below=of nmax] (problem) {problem: $i_p \in I_p = \{0,1,\ldots,8\}$};

\node[box, below=2cm of problem] (rc) {right completions $i$};
\node[box, left=of rc] (prompt) {prompt $i$};
\node[box, right=of rc] (wc) {wrong completions $i$};

\node[below=0.5cm of prompt, align=center, font=\LARGE] (prompt_eq) {$x^{i} \in \mathcal{P}$\\$i \in I \subseteq \{\mathbb{B} \times \mathbb{B} \times I_p\}$};
\node[below=0.5cm of rc, align=center, font=\LARGE] (rc_eq) {$y_w^{i,j} \in \mathcal{D}_w$\\$ \ \ \ \ j \in \mathcal{J}^i \triangleq \{0, 1, \ldots, n^i - 1\}$};
\node[below=0.5cm of wc, align=center, font=\LARGE] (wc_eq) {$y_l^{i,k} \in \mathcal{D}_l$\\$j \in \mathcal{K}^i \triangleq \{0, 1, \ldots, \bar{n}^i - 1\}$};

\draw[arrow] (root) -- (dataset);
\draw[arrow] (dataset) -- (task);
\draw[arrow] (task) -- (nmax);
\draw[arrow] (nmax) -- (problem);
\draw[arrow] (problem) -- (rc);
\draw[arrow] ($(problem.south)+(0,-0.5)$) -| (prompt);
\draw[arrow] ($(problem.south)+(0,-0.5)$) -| (wc);

\end{tikzpicture}
\caption{Schematic of the generated original dataset $\mathcal{D}^*$ disposition.}
\label{fig:streamlined-dpo-dataset-structure-trifurcation}
\end{figure*}

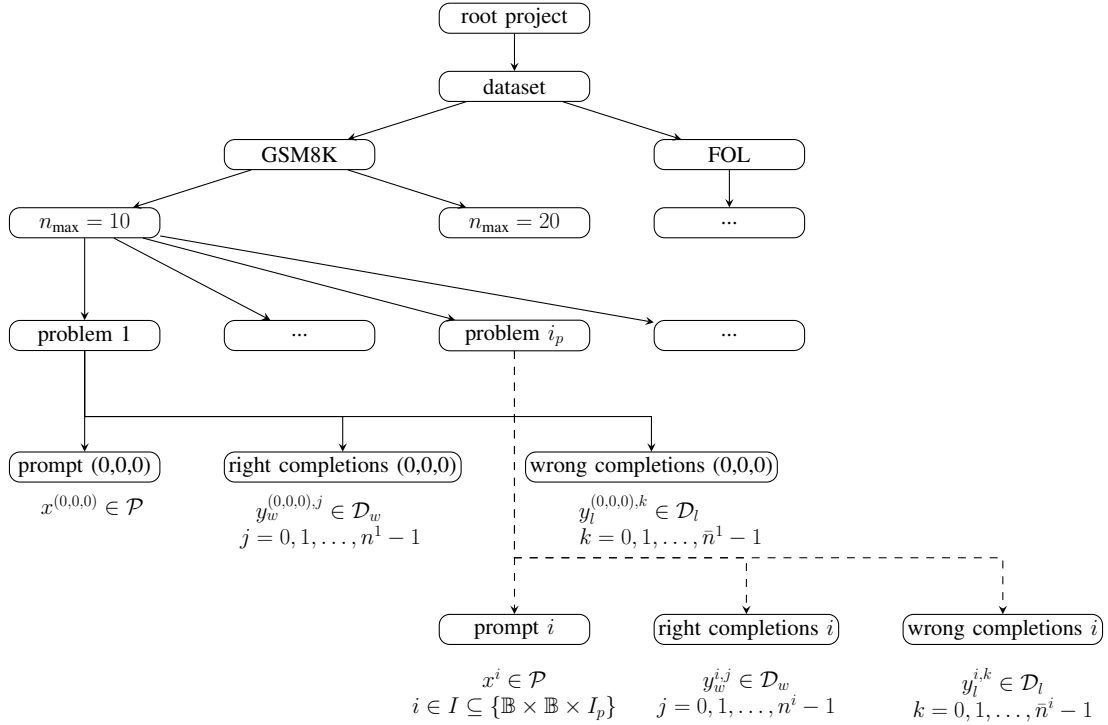
\begin{figure*}
\centering
\begin{tikzpicture}[
    scale=0.5,
    transform shape,
    node distance = 1cm and 1.7cm,
    box/.style = {rectangle, draw, rounded corners, minimum width=4cm, minimum height=0.8cm, font=\LARGE},
    arrow/.style = {->, >=stealth, thin},
    dashed arrow/.style = {->, >=stealth, thin, dashed}
]
\node[box] (root) {root project};
\node[box, below=of root] (dataset) {dataset};
\node[box, below left=of dataset] (GSM8K) {GSM8K};
\node[box, below right=of dataset] (fol) {FOL};
\node[box, below=of fol] (fol_dots) {...};
\node[box, below left=of GSM8K] (n10) {$n_{\text{max}} = 10$};
\node[box, below right=of GSM8K] (n20) {$n_{\text{max}} = 20$};
\node[box] (problem1) at ($(n10)-(0,3)$) {problem 1};
\node[box, right=of problem1] (dots1) {...};
\node[box, right=of dots1] (problemi) {problem $i_p$};
\node[box, right=of problemi] (dots2) {...};
\node[box] (prompt1) at ($(problem1)-(0,3.5)$) {prompt (0,0,0)};
\node[box, right=of prompt1] (rc1) {right completions (0,0,0)};
\node[box, right=of rc1] (wc1) {wrong completions (0,0,0)};
\node[box, below=7cm of problemi] (prompti) {prompt $i$};
\node[box, right=of prompti] (rci) {right completions $i$};
\node[box, right=of rci] (wci) {wrong completions $i$};
\node[below right=1cm and 0.3cm of prompt1.north west, align=right, font=\LARGE] (prompt_eq1) {$ \ \ x^{(0,0,0)} \in \mathcal{P}$};
\node[below right=1cm and 0.3cm of rc1.north west, align=right, font=\LARGE] (rc_eq1) {$y_w^{(0,0,0),j} \in \mathcal{D}_w \ \ \ \ \ $\\$j = 0,1,\ldots,n^{1}-1$ };
\node[below right=1cm and 1.3cm of wc1.north west, align=right, font=\LARGE] (wc_eq1) {$y_l^{(0,0,0),k} \in \mathcal{D}_l  \ \ \ \ \ \ \ \ $\\$k = 0,1,\ldots,\bar{n}^{1}-1$};
\node[below=0.5cm of prompti, align=center, font=\LARGE] (prompt_eqi) {$x^{i} \in \mathcal{P}$\\$i \in I \subseteq \{\mathbb{B} \times \mathbb{B} \times I_p\}$};
\node[below=0.5cm of rci, align=center, font=\LARGE] (rc_eqi) {$y_w^{i,j} \in \mathcal{D}_w$\\$j = 0,1,\ldots,n^{i}-1$};
\node[below=0.5cm of wci, align=center, font=\LARGE] (wc_eqi) {$y_l^{i,k} \in \mathcal{D}_l$\\$k = 0,1,\ldots,\bar{n}^{i}-1$};
\draw[arrow] (root) -- (dataset);
\draw[arrow] (dataset) -- (GSM8K);
\draw[arrow] (dataset) -- (fol);
\draw[arrow] (fol) -- (fol_dots);
\draw[arrow] (GSM8K) -- (n10);
\draw[arrow] (GSM8K) -- (n20);
\draw[arrow] (n10) -- (problem1);
\draw[arrow] (n10) -- (dots1);
\draw[arrow] (n10) -- (problemi);
\draw[arrow] (n10) -- (dots2);
\draw[arrow] (problem1) -- (prompt1);
\draw[arrow] (problem1.south) -- ($(problem1.south)+(0,-1.75)$) -| (rc1);
\draw[arrow] (problem1.south) -- ($(problem1.south)+(0,-1.75)$) -| (wc1);
\draw[dashed arrow] (problemi) -- (prompti);
\draw[dashed arrow] (problemi.south) -- ($(problemi.south)+(0,-5.5)$) -| (rci);
\draw[dashed arrow] (problemi.south) -- ($(problemi.south)+(0,-5.5)$) -| (wci);
\end{tikzpicture}
\caption{Extended schematic of the generated original dataset $\mathcal{D}^*$ disposition.}
\label{fig:complete-dpo-dataset-structure}
\end{figure*}

Our dataset $\mathcal{D}^{*} \equiv \{\mathcal{P}, \mathcal{D}_w, \mathcal{D}_l \}$ was generated using Llama3-70B through the \texttt{microchain}, producing, given a set of prompts (denoted by $\mathcal{P}$), a range of correct and incorrect chain completions (whose sets are denoted respectively by $\mathcal{D}_w$ and $\mathcal{D}_l$). The generation process encompassed various scenarios, differentiated by the type of the previous discussed reasoning tasks (\texttt{GSM8K} or FOL) and the maximum number of allowed $n$ iterations in the chain of thought ($n_{\text{max}} = 10$ or $20$), where each iteration is a couple of assistant/user role chat contents. For each problem in the dataset, we recorded a prompt, a set of correct completions (right completions), and a set of incorrect completions (wrong completions). For a better understanding of the dataset disposition, see Fig.~\ref{fig:streamlined-dpo-dataset-structure-trifurcation} and Fig.~\ref{fig:complete-dpo-dataset-structure}, where indices $i_t \in \mathbb{B}, i_n \in \mathbb{B}, i_p \in I_p \triangleq \{0,1,2,\dots,8\}$, and $i \in I \subseteq \{\mathbb{B} \times \mathbb{B} \times I_p\}, \mathbb{B} \triangleq \{0,1\}$, identify the type of task (mathematical or logic), the maximum chain iteration number (10 or 20), the problem within the specific task, ranging from 0 to 5 for the FOL problem set and 0 to 9 for the \texttt{GSM8K} problem set, and the global index, equals to $(i_t, i_n, i_p)$, respectively.
An example of a right completion $y_w^{i,j}$ for $x^i$ is shown in Tab.~\ref{tab:right_completion}, where $j \in \mathcal{J}^i \triangleq \{0, 1, \ldots, n^i - 1\}$ and $n^i$ is the number of right completions related to the $i$-th prompt. 
The function \texttt{CheckCorrectChain()} defined in the prompt, is called after that the other functions are correctly called in the right order and the \texttt{Stop()} function stops the program in $n \leq n_{max} = 10$ iterations.
Similarly, Tab~\ref{tab:right_completion_2} shows an other example $y_w^{i,j}$ for $x^i$. Although a syntax error occurs at the beginning, the chain is then correctly executed and the program stops correctly in $n \leq n_{max} = 10$ iterations.
Finally, Tab.~\ref{tab:wrong_completion} shows an example of wrong completion $y_l^{i,k}$ for $x^i$, where $k \in \mathcal{K}^i \triangleq \{0, 1, \ldots, \bar{n}^i - 1\}$ and $\bar{n}^i$ is the number of all the right completions related to the $i$-th prompt. The chain presents a syntax error and the \texttt{Stop()} function is prematurely called before it is correctly executed.
For each problem prompt, we generated $n^{i}_c=n^i+\bar{n}^i=1000$ samples for a total amount of $|\mathcal{D}^*|=15000$ samples.

\subsection{Data augmentation of the DPO dataset}

Given the original dataset $\mathcal{D}^{*}$ of all tuples $(x^{i},y_w^{i,j},y_l^{i,k})$, $\forall i \in I, \forall j \in J^i, \forall k \in K^i$, we built an augmented dataset $\mathcal{D}^a$ where for a given prompt $x^i$ a combination of all the correct completions $y_w^{i,j}$ with all the wrong completions $y_l^{i,k}$ was made, extending the cardinality of the dataset from $|\mathcal{D}^{*}| = |\mathcal{P}| = |\mathcal{D}_{w}| + |\mathcal{D}_{l}|$ to $|\mathcal{D}^{a}| = |\mathcal{D}_{w}| \times |\mathcal{D}_{l}|$. 
The final DPO dataset $\mathcal{D}$ in \eqref{eq:loss_dpo} is a subset of uniform random samples from $\mathcal{D}^{a}$, such that $\mathcal{D} \subset \mathcal{D}^a$, where the cardinality of $\mathcal{D}$ is $|\mathcal{D}| = n_s$, i.e. the number of samples belonging $\mathcal{D}$. In particular we drew a total count of $n_s=40000$ samples. This approach allows an augmentation of the original dataset. A sample of the final dataset can be written as
\begin{equation*}
( x = x^i \sim \mathcal{P}, \ y_w = y_w^{i,j} \sim \mathcal{D}_w, \ y_l = y_l^{i,k} \sim \mathcal{D}_l ) \ .
    \label{eq:sample}
\end{equation*}
We split the dataset in training and test set. Specifically, data related to 4 FOL problems and 5 \texttt{GSM8K} problems were used for the training, instead the remaining data (related to the remaining 6 problems) was used for the model inference test. Further details are given in Tab.~\ref{tab:sample_count} and Tab.~\ref{tab:sample_count_2}.

\begin{table}[ht]
\centering
\begin{tabular}{lccc}
\toprule
\textbf{Dataset} & \textbf{FOL} &  \textbf{GSM8K} & \textbf{overall} \\
\midrule
\textbf{Training set} & 17823 & 14045 & 31868\\
\textbf{Test set} & 7403 & 10729 & 18132 \\
\bottomrule
\end{tabular}
\caption{Sample count number for the train and test set, counted for the 2 different problem tasks.}
\label{tab:sample_count}
\end{table}


\begin{table}[ht]
\centering
\begin{tabular}{lcccc}
\toprule
$i_p$ & \textbf{Train (FOL)} &  \textbf{Test (FOL)} &  \textbf{Train (GSM8K)} &  \textbf{Test (GSM8K)} \\
\midrule
0 & 4421 & - & 1582 & - \\
1 & 2957 & - & 2941 & - \\
2 & 5214 & - & 2581 & - \\
3 & 5231 & - & 5039 & - \\
4 & - & 5247 & 1902 & - \\
5 & - & 2156 & - & 3381 \\
6 & - & - & - & 1673 \\
7 & - & - & - & 386 \\
8 & - & - & - & 5289 \\
\midrule
\textbf{tot.} & \textbf{17823} & \textbf{7403} & \textbf{14045} & \textbf{10729} \\
\bottomrule
\end{tabular}
\caption{Sample count number for the train and test set, counted for each of the 15 problems.}
\label{tab:sample_count_2}
\end{table}

\subsection{Inference and Training setup}

Models were deployed using the \texttt{vllm} library \cite{kwon2023efficient}. We trained the model Mistral-7B-Instruct-v0.2 on a single GPU NVIDIA A100, using the python library Transformer Reinforcement Learning \texttt{trl} \cite{vonwerra2022trl} which includes the \texttt{DPOTrainer} class, combined with the use of other three libraries specific for the training optimization process, namely \texttt{accelerate} \cite{accelerate}, \texttt{deepspeed} \cite{rasley2020deepspeed} and \texttt{peft} \cite{peft}. Training parameters used for DPO are specified in Tab.~\ref{tab:dpo}.
Tab.~\ref{tab:machine} and \ref{tab:soft_environment} summarize the hardware and the software specification used, respectively. Note that despite machine specifications report 4 GPU, only one GPU was used for the training process.    


\begin{table}[htbp]
\centering
\begin{tabular}{|p{0.42\textwidth}|}
\hline
\textbf{Prompt} \\
\hline
\begin{minipage}[t]{0.93\textwidth}
\small
\begin{verbatim}

Act as a movie expert. 
You can use the following functions:

Reasoning(reasoning: str)
Use this function for your internal 
reasoning.
Example: 
Reasoning(reasoning="The next step to 
take is...")

Actor(name: str)
Predicate to check if a given name 
is an actor.
Example: 
Actor(name="Sean Connery")

Movie(x: str)
Predicate that queries IMDb to determine
if the argument is a movie.
Example: 
Movie(x="Goldfinger")

ActsIn(actor: str, movie_title: str)
Check if a specific actor acted in 
a given movie.
Example: 
ActsIn(actor="Sean Connery", 
movie_title="Goldfinger")

CheckCorrectChain()
Check if the labels are correct.
Example: 
CheckCorrectChain()

Stop()
Use this function to stop the program.
Example: 
Stop()

Verify if Tom Hanks acted in 
the movie Cast Away.

\end{verbatim}
\end{minipage} \\
\hline
\end{tabular}
\caption{An example of a prompt $x^i$ of a sample of the dataset $\mathcal{D^*}$.}
\label{tab:movie_expert_prompt}
\end{table}

\begin{table*}[htbp]
\centering
\begin{tabular}{|p{1.01\textwidth}|}
\hline
\textbf{Right Completion} \\
\hline
\begin{minipage}[t]{0.93\textwidth}
\small
\begin{verbatim}

{"role": "assistant", "content": 
"Reasoning("I need to reason step-by-step, checking if the actor and the film actually exist")"}
{"role": "user", "content": "The reasoning has been recorded"}
{"role": "assistant", "content": 
"Reasoning("First, I'll check if Tom Hanks is an actor")"}
{"role": "user", "content": "The reasoning has been recorded"}
{"role": "assistant", "content": "Actor(name="Tom Hanks")"}
{"role": "user", "content": "True"}
{"role": "assistant", "content": 
"Reasoning("Next, I'll check if Cast Away is a movie")"}
{"role": "user", "content": "The reasoning has been recorded"}
{"role": "assistant", "content": "Movie(x="Cast Away")"}
{"role": "user", "content": "True"}
{"role": "assistant", "content": 
"Reasoning("Now, I'll check if Tom Hanks acted in Cast Away")"}
{"role": "user", "content": "The reasoning has been recorded"}
{"role": "assistant", "content": "ActsIn(actor="Tom Hanks", movie_title"Cast Away")"}
{"role": "user", "content": "True"}
{"role": "assistant", "content": 
"Reasoning("Based on the information obtained, Tom Hanks did indeed act in Cast Away")"}
{"role": "user", "content": "The reasoning has been recorded"}
{"role": "assistant", "content": "CheckCorrectChain()"}
{"role": "user", "content": "True"}
{"role": "assistant", "content": "Stop()"}
{"role": "user", "content": "The program has been stopped"}

\end{verbatim}
\end{minipage} \\
\hline
\end{tabular}

\caption{An example of a right completion $y_w^{i,j}$ of a sample of the dataset $\mathcal{D}^*$. The chain presents the functions called in the correct order.}

\label{tab:right_completion}
\end{table*}

\begin{table*}[htbp]
\centering
\begin{tabular}{|p{1.01\textwidth}|}
\hline
\textbf{Right Completion} \\
\hline
\begin{minipage}[t]{0.93\textwidth}
\small
\begin{Verbatim}[commandchars=\\\{\}]

\{"role": "assistant", "content": 
"Reasoning("I need to reason step-by-step, checking if the actor and the film actually exist")"\}
\{"role": "user", "content": "The reasoning has been recorded"\}
\textcolor{red}{\{"role": "assistant", "content":} 
\textcolor{red}{"Reasoning(reasoning=Check if Tom Hanks is an actor)"\}}
\textcolor{red}{\{"role": "user", "content":}
\textcolor{red}{"Error: syntax error in command} 
\textcolor{red}{Reasoning(reasoning=Check if Tom Hanks is an actor). Please try again."\}}
\{"role": "assistant", "content": "Actor(name="Tom Hanks")"\}
\{"role": "user", "content": "True"\}
\{"role": "assistant", "content": "Movie(x="Cast Away")"\}
\{"role": "user", "content": "True"\}
\{"role": "assistant", "content": "ActsIn(actor="Tom Hanks", movie_title="Cast Away")"\}
\{"role": "user", "content": "True"\}
\{"role": "assistant", "content": 
"Reasoning("All the checks have been successful, 
I can conclude that Tom Hanks acted in Cast Away")"\}
\{"role": "user", "content": "The reasoning has been recorded"\}
\{"role": "assistant", "content": "CheckCorrectChain()"\}
\{"role": "user", "content": "True"\}
\{"role": "assistant", "content": "Stop()"\}
\{"role": "user", "content": "The program has been stopped"\}

\end{Verbatim}
\end{minipage} \\
\hline
\end{tabular}

\caption{An example of a right completion $y_w^{i,j}$ of a sample of the dataset $\mathcal{D}^*$. At the beginning the chain presents a syntax error in the function calling, but then the completion is correctly executed.}

\label{tab:right_completion_2}
\end{table*}

\begin{table*}[htbp]
\centering
\begin{tabular}{|p{1.01\textwidth}|}
\hline
\textbf{Wrong Completion} \\
\hline
\begin{minipage}[t]{0.93\textwidth}
\small
\begin{Verbatim}[commandchars=\\\{\}]

\{"role": "assistant", "content": 
"Reasoning("I need to reason step-by-step, checking if the actor and the film actualy exist")"\}
\{"role": "user", "content": "The reasoning has been recorded"\}
\textcolor{red}{\{"role": "assistant", "content":} 
\textcolor{red}{"Reasoning(reasoning=Check if Tom Hanks is an actor)"\}}
\textcolor{red}{\{"role": "user", "content":}
\textcolor{red}{"Error: syntax error in command} 
\textcolor{red}{Reasoning(reasoning=Check if Tom Hanks is an actor). Please try again."\}}
\{"role": "assistant", "content": "CheckCorrectChain()"\}
\{"role": "user", "content": "False"\}
\{"role": "assistant", "content": "Stop()"\}
\{"role": "user", "content": "The program has been stopped"\}

\end{Verbatim}
\end{minipage} \\
\hline
\end{tabular}

\caption{An example of a wrong completion $y_l^{i,k}$ of a sample of the dataset $\mathcal{D}^*$. The chain presents a syntax error, and the function \texttt{Stop()} is called before the chain is correctly executed.}

\label{tab:wrong_completion}
\end{table*}


        \begin{table}[ht]
            \centering
            \scalebox{1.6}{
            \begin{tabular}{ll}
            \hline
            \textbf{Parameter} & \textbf{Value} \\
            \hline
            \hline
            Sequence Length & 3072 \\
            Micro Batch Size & 1 \\
            Number of Epochs & 3 \\
            Optimizer & \texttt{adamw\_bnb\_8bit} \\
            Learning Rate & \texttt{5e-8} (with decay) \\
            LR Scheduler & cosine \\
            Zero Stage & 3 \\
            Warmup Steps & 10 \\
            \hline
            \end{tabular}
            }
            \caption{ \vspace{0.1cm} Parameter for model training in DPO.}
            \label{tab:dpo}
        \end{table}

        \begin{table}[ht]
            \centering
            \scalebox{1.15}{
            \begin{tabular}{ll}
                \hline
                Machine & DELL xe8545\\
                CPU & 96 x AMD EPYC 7413, 24Core@2.5GHz  \\
                RAM & 32 x DDR4 32GB \\
                GPU & 4 x A100-SXM4 (80GB) with NVIDIA NVLink \\
                \hline
            \end{tabular}
            }
            \caption{Machine specifications.}
            \label{tab:machine}
        \end{table}
    
        \begin{table}[ht]
         \centering
         \scalebox{1.6}{
            \begin{tabular}{ll}
                \hline
            OS & Linux Ubuntu 20.04.6 LTS \\
            Python & v3.8.10\\
            torch & v2.3.0\\
            transformers & v4.42.3 \\
            trl & v0.8.6 \\
            accelerate & v0.31.0 \\
            deepspeed & v0.14.2 \\
            peft & v0.11.1 \\
            wandb & v0.17.0 \\
            \hline
            \end{tabular}
            }
            \caption{Software environment.}
            \label{tab:soft_environment}
        \end{table}


\subsection{Performance metrics}
\label{subsec:metrics}

After the training of the small-scale LLM, we tested performance in inference of both the original and trained small-scale LLM, producing 2 new different datasets $\mathcal{D}^*_O$ and $\mathcal{D}^*_T$, respectively, with $1000$ generated samples for each task problem. Metric used to compare model performances is the accuracy, defined for the $i$-th problem prompt as $a^i\triangleq|\mathcal{J}^i|/(|\mathcal{J}^i|+|\mathcal{K}^i|)=n^i/n^{i}_{c}$. The task average accuracy is defined as $\bar{a}_{i_t} \triangleq \sum_{i \in I_t} a^i / |I_t|$ where $I_t \subset I$ is the problem index subset related to the $t$-th task. The average accuracy computed on subsets of $I_t$ related to different $n_{max}$ values and types of data subsets (training and test type of data subsets), for the sake of brevity is still denoted with $\bar{a}_{i_t}$. Finally, the overall average accuracy is defined as $\bar{a} \triangleq \sum_{i \in I} a^i / |I|$.

\section{Experimental results}
\label{sec:exps}

\begin{figure}[htbp]   
    \centering
    \includegraphics[width=1\linewidth]{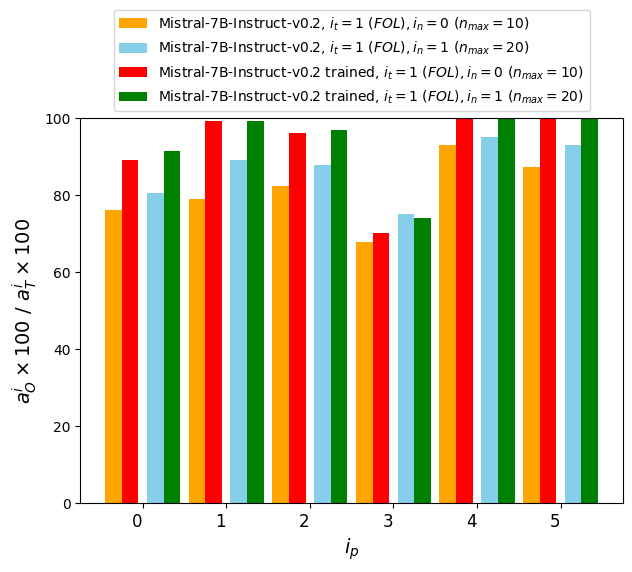}
    \caption{Comparison of both the original and the trained small-scale model percentage accuracies $a^i\times100$, for different FOL data subset problems and $n_{max}$ values.}
    \label{fig:fol_accuracy}
\end{figure}

\begin{figure}[htbp]   
    \centering
    \includegraphics[width=1\linewidth]{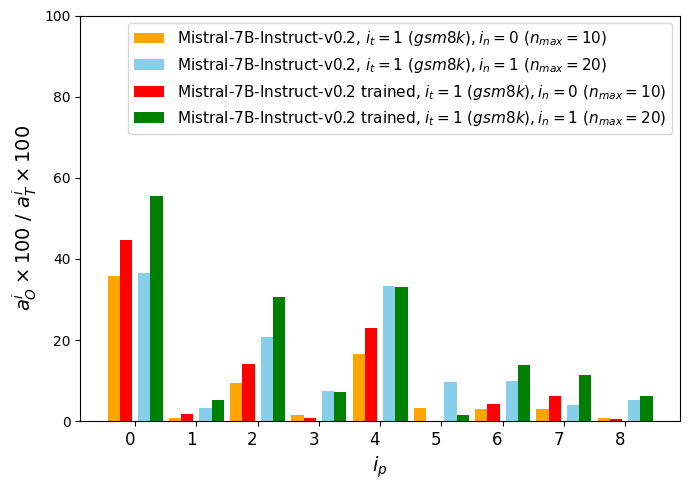}
    \caption{Comparison of both the original and the trained small-scale model percentage accuracies $a^i\times100$, for different \texttt{GSM8K} data subset problems and $n_{max}$ values.}
    \label{fig:gsm8k_accuracy}
\end{figure}

\begin{figure}[htbp]   
    \centering
    \includegraphics[width=1\linewidth]{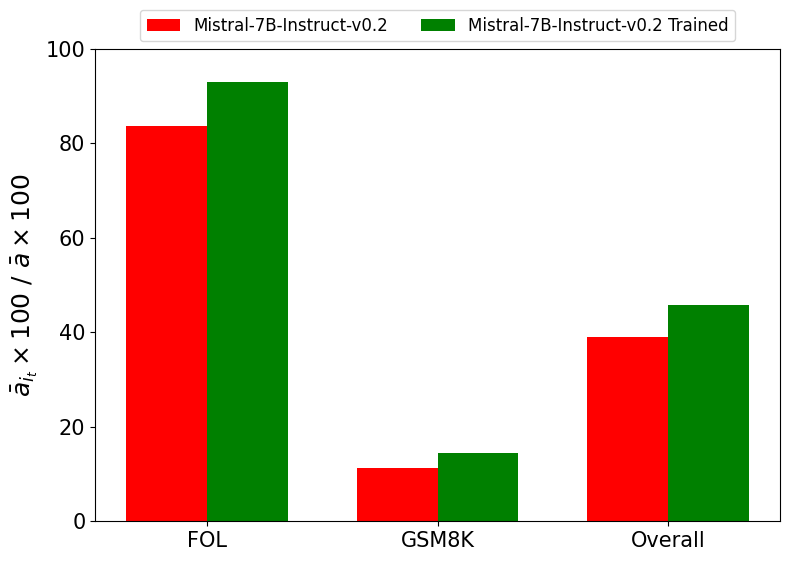}
    \caption{Comparison of the task average percentage accuracies $\bar{a}_{i_t}\times100$ and the overall average percentage accuracy $\bar{a}\times100$ calculated for the original and the fine-tuned model, for different task data subsets and the whole dataset.}
    \label{fig:overall}
\end{figure}

\begin{figure}[htbp]   
    \centering
    \includegraphics[width=1\linewidth]{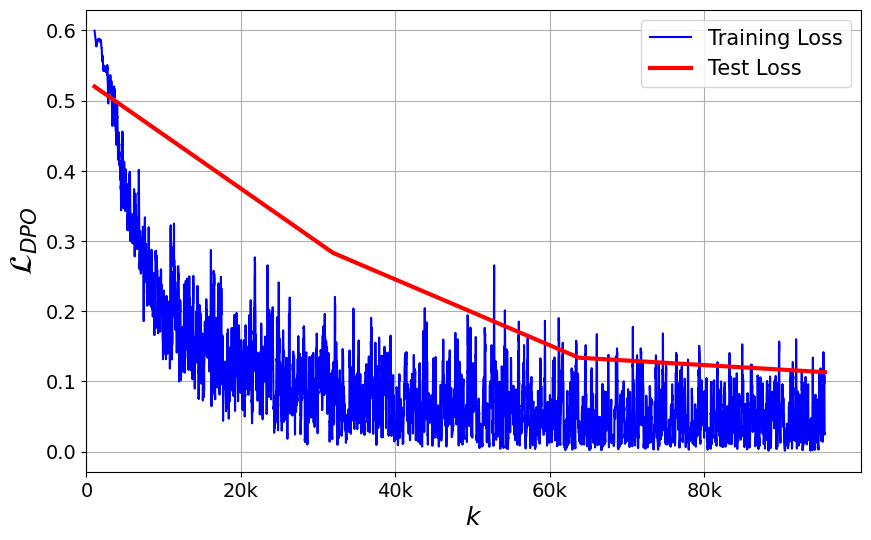}
    \caption{Training loss vs test loss over global step $k$.}
    \label{fig:loss}
\end{figure}

The percentage accuracy $a^i_O\times100$ for the problem $i$-th and the dataset $\mathcal{D}_O^*$, is compared to $a^i_T\times100$ for the dataset $\mathcal{D}_T^*$, as shown in Fig.~\ref{fig:fol_accuracy} and \ref{fig:gsm8k_accuracy}. We can see how FOL task performance is completely improved, except for the problem 3 with $n_{max} = 20$. Model performance is also improved on the \texttt{GSM8K} data subset, where on the whole is better than the original model, apart from some cases. 
Note that, as expected, performance with $n_{max} = 20$ is better than one with $n_{max} = 10$, for both the original and the fine-tuned models. 
Overall performance is so improved, but the fine-tuned model struggles more with the \texttt{GSM8K} problems. The same trend is observed in Tab.~\ref{tab:accuracy}, where task average percentage accuracy $\bar{a}_{i_t}\times100$ and overall average percentage accuracy $\bar{a}\times100$ evaluations are reported for different models, types of data subsets and $n_{max}$ values. Note that also performances on the training and test set, taken individually, are improved after the fine-tuning process for all the configurations of $n_{max}$ values and tasks. The trend is also confirmed in Fig.~\ref{fig:overall}, where a comparison of the original and the fine-tuned model task average percentage accuracies $\bar{a}_{i_t}\times100$ with the overall percentage accuracy $\bar{a}_{i_t}\times100$ is presented. Furthermore, results are statistically evaluated with a Wilcoxon signed-rank test where $p$-values are computed comparing the original and the fine-tuned model accuracies, for each task data subset and the whole dataset. Values are all lower than $0.05$, confirming that the difference is statistically significant, as shown in Tab.~\ref{tab:wilcoxon_results}, where $W$ represents the Wilcoxon signed-rank test statistic. Finally, loss metric in \eqref{eq:loss_dpo} computed at each global step $k$ during the training process is shown in Fig.~\ref{fig:loss}, where the evaluation on the test set is executed at the end of each epoch. A global step represents a single update of the model parameters. It is incremented every time the optimizer performs a backpropagation operation and updates the model weights. Each epoch implies that the model has processed all the preference pairs (right and wrong completions) present in the dataset. 
The loss metric is computed using the \texttt{wandb} library \cite{wandb}.

\begin{table}[ht]
\centering
\begin{tabular}{lcccc}
\toprule
$n_{max}=10$ & & & & \\
\midrule
\textbf{Model} & \textbf{Dataset} & \textbf{FOL} &  \textbf{GSM8K} & \textbf{Overall} \\
\midrule
\textbf{original} & \textbf{Training set} & 76.47\% & 12.82\% & 44.65\% \\
\textbf{original} & \textbf{Test set} & 90.19\% & 2.56\% & 46.38\% \\
\textbf{original} & \textbf{Whole set} & 81.04\% & 8.26\% & 44.65\% \\
\textbf{fine-tuned} & \textbf{Training set} & \textbf{88.75\%} & \textbf{16.81\%} & \textbf{52.78\%} \\
\textbf{fine-tuned} & \textbf{Test set} & \textbf{99.75\%} & \textbf{2.77\%} & \textbf{51.26\%} \\
\textbf{fine-tuned} & \textbf{Whole set} & \textbf{92.42\%} & \textbf{10.57\%} & \textbf{51.50\%} \\
\midrule
\midrule
$n_{max}=20$ & & & & \\
\midrule
\textbf{Model} & \textbf{Dataset} & \textbf{FOL} &  \textbf{GSM8K} & \textbf{Overall} \\
\midrule
\textbf{original} & \textbf{Training set} & 83.18\% & 20.22\% & 51.70\% \\
\textbf{original} & \textbf{Test set} & 94.14\% & 7.22\% & 50.68\% \\
\textbf{original} & \textbf{Whole set} & 86.84\% & 14.45\% & 50.65\% \\
\textbf{fine-tuned} & \textbf{Training set} & \textbf{90.50\%} & \textbf{26.36\%} & \textbf{58.43\%} \\
\textbf{fine-tuned} & \textbf{Test set} & \textbf{99.75\%} & \textbf{8.18\%} & \textbf{53.97\%} \\
\textbf{fine-tuned} & \textbf{Whole set} & \textbf{93.58\%} & \textbf{18.28\%} & \textbf{55.93\%} \\
\midrule
\midrule
$n_{max} = 10$ and $20$ & & & & \\
\midrule
\textbf{Model} & \textbf{Dataset} & \textbf{FOL} &  \textbf{GSM8K} & \textbf{Overall} \\
\midrule
\textbf{original} & \textbf{Training set} & 79.83\% & 16.52\% & 48.18\% \\
\textbf{original} & \textbf{Test set} & 92.16\% & 4.89\% & 48.53\% \\
\textbf{original} & \textbf{Whole set} & 83.94\% & 11.35\% & 47.65\% \\
\textbf{fine-tuned} & \textbf{Training set} & \textbf{89.62\%} & \textbf{21.58\%} & \textbf{55.60\%} \\
\textbf{fine-tuned} & \textbf{Test set} & \textbf{99.75\%} & \textbf{5.47\%} & \textbf{52.61\%} \\
\textbf{fine-tuned} & \textbf{Whole set} & \textbf{93.00\%} & \textbf{14.42\%} & \textbf{53.71\%} \\
\bottomrule
\end{tabular}
\caption{Task average percentage accuracy $\bar{a}_{i_t}$ and overall average percentage accuracy $\bar{a}\times100$ evaluations of the original and fine-tuned model for different datasets and $n_{max}$ values.}
\label{tab:accuracy}
\end{table}

\begin{table}[h]
    \centering
    \begin{tabular}{lcc}
        \hline
        \textbf{Dataset} & \textbf{$W$} & {$p$-value} \\
        \hline
        FOL     & 77  & \textbf{4.88e-4} $< \alpha$\\
        GSM8K   & 137 & \textbf{1.18e-2} $< \alpha$\\
        Overall & 416 & \textbf{2.50e-5} $< \alpha$\\
        \hline
    \end{tabular}
    \caption{Wilcoxon signed-rank test results comparing original and fine-tuned model accuracies for different datasets. Significant $p$-values $(p < \alpha = 0.05)$ are highlighted in bold. $W$ is the Wilcoxon test statistic.}
    \label{tab:wilcoxon_results}
\end{table}



\section{Conclusions and future works}

In this study, we introduced a novel framework for improving the function calling abilities of small-scale LLMs, focusing on specific logical and mathematical reasoning tasks. Our approach addresses the inefficiencies and high computational costs associated with relying solely on large-scale LLMs by leveraging the capabilities of small-scale models through RLHF.
We employed an agent-based system that interacts with a large-scale LLM to generate a dataset comprising correct and incorrect step-by-step reasoning chain chat completions in the domains of First-Order Logic (FOL) and mathematical reasoning tasks drawn from the \texttt{GSM8K} dataset. Utilizing this dataset, we trained a smaller model, Mistral-7B-Instruct-v0.2, employing RLHF with the DPO technique.
Our experimental results demonstrate significant improvements in the performance of the small-scale model on FOL tasks, achieving near-perfect accuracy in most cases. While the improvements on the \texttt{GSM8K} mathematical problems are more modest, the trained model still outperforms the original model in overall accuracy. These findings suggest that our framework effectively improves the function calling abilities of smaller models enhancing their capabilities in the using of external tools (the callable functions) and the abilities in the given reasoning tasks.
By successfully improving the integration of small-scale LLMs with external function calls, our approach contributes to making advanced reasoning tasks more accessible and practical in resource-constrained environments, since the training process was carried out by using just a single GPU. This has significant implications for deploying AI systems on devices with limited computational resources, such as mobile devices or real-time applications.
For future work, we plan to explore the application of our framework to a broader range of reasoning tasks and function types. 
Indeed, expanding the methodology to include more diverse datasets and functions could help in testing the generalization of the approach and increasing its applicability across various domains.
Additionally, we aim to investigate further optimizations in the training process, for instance through a hyperparameter optimization process \cite{yang2020hyperparameter, manduzio2024advanced}, by varying values of the model training parameters presented in Tab.~\ref{tab:dpo}. 
One other promising area for optimization that we plan to investigate is the use of quantization to reduce the size of the model and improve inference speed \cite{jacob2018quantization}. Quantization reduces the precision of model parameters from floating-point (e.g., 32-bit) to lower bit-width representations (e.g., 8-bit or even binary), allowing models to operate more efficiently, especially on hardware with limited resources.
Knowledge distillation is another key technique that can further enhance the performance of small-scale models by transferring knowledge from larger models \cite{hinton2015distilling}. In this approach, a large teacher model, which has learned complex reasoning patterns, guides a smaller student model to replicate its behavior. This process allows the smaller model to maintain high performance despite its reduced size. For future works, we plan to investigate whether knowledge distillation can be combined with the DPO algorithm in the proposed framework.
Furthermore, we plan to comprehend whether the proposed approach can also enhance the intrinsic reasoning capabilities of LLMs by enabling them to solve problems without relying on external functions and tools.


\section*{Acknowledgement}
The research leading to these results has received partial funding from the Italian Ministry of Education and Research (MIUR) in the framework of the ForeLab project (Departments of Excellence), from PNRR - M4C2 - Investimento 1.3, Partenariato Esteso PE00000013- “FAIR- Future Artificial Intelligence Research” - Spoke 1 “Human-centered AI”, funded by the European Commission under the NextGeneration EU programme, from the PRIN grant no. 2022ALBSWX of the Italian Ministry of University and Research, and partial funding from the Italian Ministry of Enterprises and Made in Italy, in the framework of the "Agreements for Innovation" Project "4DDS - 4D Drone Swarms" Ref. no. F/310097/01-04/X56.

\FloatBarrier

\bibliographystyle{IEEEtran}
\bibliography{main}  

\end{document}